# Is the Fitness Dependent Optimizer Ready for the Future of Optimization?

Ardalan H. Awlla
Information Technology, Sulaimani Polytechnic University, Sulaymaniyah 46001, Iraq
ardalan.husin.tci@spu.edu.iq
Tarik A. Rashid
Computer Science and Engineering Department; Centre for Artificial Intelligence and Innovations, University of Kurdistan Hewler, Erbil 44001, Iraq
tarik.ahmed@ukh.edu.krd
Ronak M. Abdullah
Department of Mathematics, College of Science, University of Sulaimani, Kurdistan Region, Iraq
runak.abdullah@univsul.edu.iq

## ABSTRACT

Metaheuristic algorithms are optimization methods that are inspired by real phenomena in nature or the behavior of living beings, e.g., animals, to be used for solving complex problems, as in engineering, energy optimization, health care, etc. One of them was the creation of the Fitness Dependent Optimizer (FDO) in 2019, which is based on bee-inspired swarm intelligence and provides efficient optimization. This paper aims to introduce a comprehensive review of FDO, including its basic concepts, main variations, and applications from the beginning. It systematically gathers and examines every relevant paper, providing significant insights into the algorithm's pros and cons. The objective is to assess FDO's performance in several dimensions and to identify its strengths and weaknesses. This study uses a comparative analysis to show how well FDO and its variations work at solving real-world optimization problems, which helps us understand what they can do. Finally, this paper proposes future research directions that can help researchers further enhance the performance of FDO.

## 1. Introduction

Metaheuristic algorithms are powerful optimization approaches which use effective strategies to improve heuristic methods' efficiency in identifying the optimal solution for challenging problems. Today, metaheuristic algorithms are widely applied to solve difficult optimization problems in diverse sectors such as engineering, economics, and logistics [1]. This makes them appeal for scenarios where more traditional optimization methods are challenged, like gradient-based algorithms, for example, when their use of gradient information is hindered, or they have to search for large solution spaces that may be complex and even multimodal. This is due to their deterministic nature, which can lead them to become stuck in local optima as shown in Figure 1, global solution refers to finding a global solution and avoiding local optima.

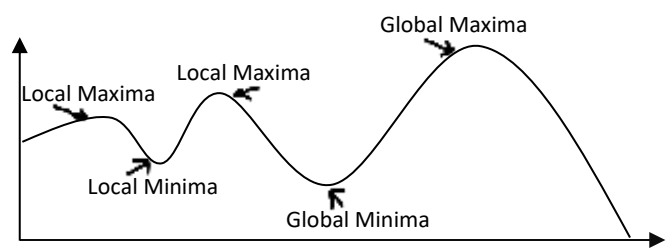

Figure 1. Local vs Global optima solution

To address these challenges, meta-heuristic-like (swarm and evolutionary) algorithms were introduced. These algorithms use stochastic processes and models inspired by biological systems to enable more effective global search capabilities [2]. Some important characteristics of metaheuristics are facilitating exploration (searching unvisited regions that contain more optimal solutions) and exploitation (moving search agents to areas along trajectories toward locally optimal regions). These algorithms, designed based on natural processes like evolution, swarm behavior, and physical phenomena, can be broadly divided into five classes: evolutionary-based, trajectory-based, art-inspired, ancient-inspired, and nature-inspired as shown in Figure 2. Evolutionary-based algorithms simulate the process of natural evolution to create high or nearly high-quality solutions for complex problems. These algorithms make use of a population of candidate solutions that evolve through generations according to operators like selection, crossover, and mutation, some of the well-known evolutionary algorithms are genetic algorithm (GA) [3], memetic algorithm (MA) [4]. Trajectory-based algorithms are iterative improvement processes, but they develop a single solution to iteratively move toward the optimum by moving through its neighborhood. They typically strike a balance of intensification and diversification to avoid local optima, among the most well-known trajectory algorithms are Simulated Annealing (SA) [5], Guided Local Search [6] and Iterative Local Search [7]. Art algorithms inspired by art base their finding of optimization on innovating through principles in creative work for instance Color Harmony Algorithm (CHA) [8] and Stochastic Paint Optimizer (SPO) [9]. Algorithms with ancient inspiration replicate elements from ancient architecture, engineering, and culture into contemporary optimization problems for example Giza Pyramids Construction (GPC) [10] and Great Wall Construction Algorithm (GWCA) [11].

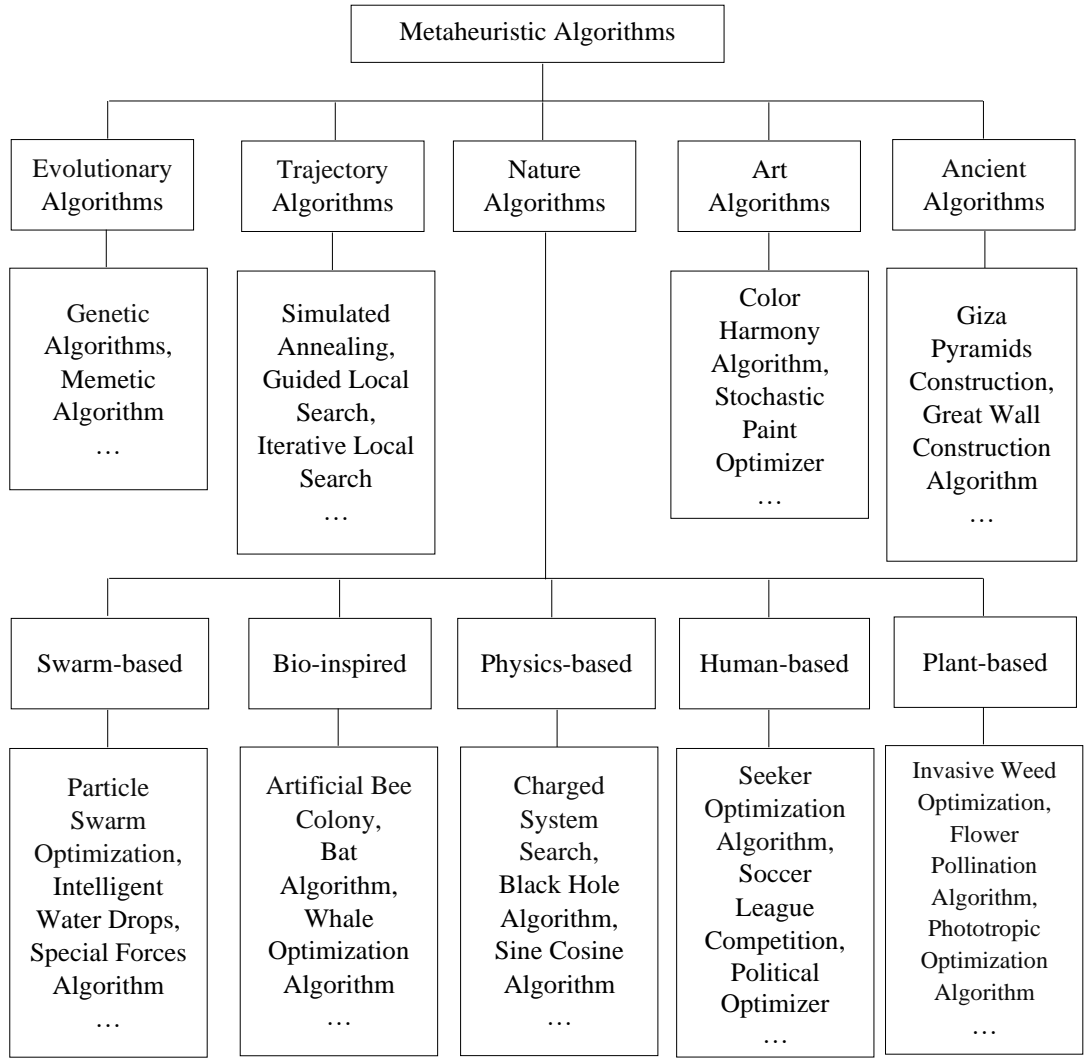

Figure 2. Classification of Metaheuristic Algorithms

Nature-inspired algorithms based on behavioral patterns in nature and physical/chemical phenomena are widely used to address various optimization tasks because of their capacity for exploration and exploitation [12]. Of these, optimization based on Swarm is the most common, and Particle Swarm Optimization (PSO) is among the most commonly used methods. PSO mimics the social behavior exhibited by birds flocking together and guides particles through the solution space while searching for optimal solutions. Because of this simplicity and success in moving the process, PSO has become a classic algorithm in swarm optimization [13], but also other algorithms like Ant Colony Optimization (ACO) [14] and Firefly Algorithm (FA) [15] are well-known.

Jaza and Tarik [16] introduced FDO, a swarm algorithm based on the reproductive behavior and collective decision-making of swarms of bees. FDO is unique among bee-inspired algorithms, like the Artificial Bee Colony (ABC) [17], for its attention to the modeling of scouts. Although the proposed algorithm draws inspiration from PSO, it employs a distinct method to adjust the agents' positions. The FDO algorithm adopts a weight mechanism based on fitness ($fw$), allowing it to have a stronger capability in balancing exploration and exploitation. Despite encountering some local optima, the FDO achieves faster convergence, enabling it to outperform other metaheuristic algorithms in practical real-world scenarios. At a minimum, it has demonstrated competitive performance with different methods, significantly improving local optima on various real-world problems, IoT Healthcare [18], and Proof Searching in HOL4 [19].

This survey's primary contribution is to collect all papers related to FDO and summarize its framework, mathematical formulations, and hybrid extensions. We discuss FDO mechanisms, like the calculation of fitness-weight and position updates. We cover different variants of FDO (e.g., Improved and Multi-Objective FDO variants) with relevant advantages and disadvantages. In addition, we explore FDO's wide applicability in fields like engineering optimization, machine learning, and data analytics, demonstrating its versatility in addressing complex problems. The survey also highlights potential future work.

We structure this paper as follows: Section 2: We highlight the basics of FDO. Section 3 discusses variants of the FDO, their respective advantages, and limitations. Applications of the FDO are presented in Section 4. Section 5 highlights the challenges and potential directions for future research. Section 7 provides the conclusion.

## 2. Basic of FDO

As mentioned in the previous section, FDO is a recent intelligent swarm optimization algorithm proposed in 2019, drawing inspiration from the search behavior of bee swarms and their reproduction procedure, which focuses on locating the best hives. The FDO comprises two primary components: the scout bee exploration process and the scout bee transition process. In the searching phase, the algorithm instructs scout bees to investigate multiple potential hives (solutions) to identify the most optimal one. During the movement phase, the algorithm uses a random walk and a fitness weight technique to adjust the positions of scout bees, facilitating their relocation to new areas. This section consists of two fundamental components.

### A. Search (Exploration)

In this process scout bees explore numerous potential hives (solution) to identify the best among them. In FDO, the scout bee population is initialized randomly within the search space. Each scout bee's position and the fitness function determine where its hives are located. Using their ability to explore nearby areas, scout bees aim to find a superior hive (new solution). If the new solution improves upon the previous one, the scout bee discards the earlier solution. However, if the scout bee finds no better solution, it adjusts its position based on the prior solution. Each scout bee is defined as $x_i$ (where $i = 1, 2, ..., n$).

### B. Movement Process of the Scout Bee

Scout bees, using fitness weight and random walk strategies, can conduct random searches within the search space. Equation 1 states that the rate at which a scout bee's current position changes influence its movement. By adjusting this pace, the scout seeks to explore a solution that is better than its previous findings.

$$X_{i,t+1} = X_{i,t} + pace \tag{1}$$

In this equation 1, $X$ represents the artificial scout bee, where $i$ stands for the individual scout bee serving as the search agent, and $t$ signifies the current iteration. The movement of the scout bee, defined by its pace, is determined by the fitness weight ($fw$). A randomization process fundamentally drives this pace. Therefore, the formula of $fw$ can be expressed as follows:

$$fw = \left| \frac{X^*_{i,t\,fitness}}{X_{i,t\,fitness}} \right| - wf \tag{2}$$

In this equation 2, $X^*_{i,t\,fitness}$ represents the most optimal global solution up to now identified by scout bee. Meanwhile, $X_{i,t\,fitness}$ represents the current solution. The weight factor ($wf$), is a randomly chosen value between 0 and 1, serving to control the $fw$. The algorithm subsequently evaluates conditions for $fw$. If $fw$ is equal to 1 or 0 and the fitness $X_{i,t\,fitness}$ is equal to 0, the pace is determined randomly according to equation (3). If $fw$ is between 0 and 1 ($fw > 0$ and $fw < 1$), the algorithm produces a random value between -1 and 1 means [1, -1], enabling the scout to explore various directions. Equation (4) computes the pace for generated values of $r$ that are $< 0$, whereas equation (5) establishes the pace for $r$ values that $\geq 1$.

$$\begin{cases} fw = 1 \ or \ fw = 0 \ or \ X_{i,t \ fitness} = 0, & pace = X_{i,t} * r & (3) \\ fw > 0 \ and \ fw < 1 \ \begin{cases} r < 0, pace = \left(X_{i,t} - X_{i,t}^*\right) * fw * -1 & (4) \\ r \geq 0, pace = \left(X_{i,t} - X_{i,t}^*\right) * fw & (5) \end{cases} \end{cases}$$

where $r$ is a randomly chosen, value within the interval $[-1,1]$, $X_{i,t}$ represents the present solution, and $X_{i,t}^*$ signifies the best global solution attained up to this point. By adjusting its $fw$ calculation and solution-selection mechanisms, the FDO effectively addresses both minimization and maximization problems. In minimization tasks, the goal is to reduce the objective function's value, while in maximization problems, it aims to increase it.

| No. | Algorithm 1 Pseudocode for the FDO algorithm |
|---|---|
| 1 | Initialize scout bee population $X_{i,t}$ (i = 1, 2, ..., n) |
| 2 | **while** iteration (t) limit not reached |
| 3 |   **for** each artificial scout bee $X_{i,t}$ |
| 4 |     find the best artificial scout bee $X_{i,t}^*$ |
| 5 |     generate a random walk $r$ in the range [-1, 1] |
| 6 |     **if** $(X_{i,t \ fitness} == 0)$ then |
| 7 |       set fitness weight = 0 |
| 8 |     **else** |
| 9 |       calculate fitness weight using equation (2) |
| 10 |     **end if** |
| 11 |     **if** (fitness weight == 1 or fitness weight == 0) then |
| 12 |       calculate pace using equation (3) |
| 13 |     **else** |
| 14 |       generate a random number |
| 15 |       **if** (random number >= 0) then |
| 16 |         calculate pace using equation (5) |
| 17 |       **else** |
| 18 |         calculate pace using equation (4) |
| 19 |       **end if** |
| 20 |     **end if** |
| 21 |     calculate new position $X_{i+1,t}$ using equation (1) |
| 22 |     **if** $(X_{i+1,t \ fitness} < X_{i,t \ fitness})$ then |
| 23 |       accept the move and save pace |
| 24 |     **else** |
| 25 |       recalculate new position $X_{i+1,t}$ using previous pace (equation 1) |
| 26 |       **if** $(X_{i+1,t \ fitness} < X_{i,t \ fitness})$ then |
| 27 |         accept the move and save pace |
| 28 |       **else** |
| 29 |         maintain current position (don't move) |
| 30 |       **end if** |
| 31 |     **end if** |
| 32 |   **end for** |
| 33 | **end while** |

## 3. Variants of FDO

Since the FDO was initially suggested, various changes and adjustments have been introduced to improve its performance and make it suitable for a wider range of optimization problems. Many of these modifications aim to tackle the limitations of the original FDO, like slow convergence, and expand its use by combining it with other optimization techniques. This section looks at the main versions of FDO, explaining their unique novelty, benefits, and drawbacks. A summary of this information is provided in Table 5.

## 3.1 Modification of Fitness Dependent Optimizer

The FDO algorithm is known for being effective at finding optimal solutions, but it has several limitations, such as slow convergence and an imbalance between exploration and exploitation. Parameters such as $wf$ and pace, which impact the algorithm's convergence, link to these issues. Setting $wf$ to zero negatively affects the speed of convergence. Additionally, the algorithm finds it challenging to balance exploration and exploitation due to its reliance on $wf$, the current fitness, and the best agent's fitness. Randomized pace settings further worsen this imbalance, causing the algorithm to perform less effectively than other algorithms. Its solutions often become suboptimal, leading to a failure to fully utilize its potential. This paper [20] introduced improved fitness-dependent optimizer (IFDO) to address exploration and exploitation issues and improve search efficiency and precision. IFDO incorporates alignment and cohesion behaviors during scout position updates, enabling better group motion and improved search capabilities. As shown in equation 6.

$$X_{i,t+1} = X_{i,t} + pace + (alignment * \frac{1}{cohesion}) \tag{6}$$

Where, alignment is when a scout's pace matches that of other scouts in its area, and cohesion is when scouts move toward the neighborhood's center of mass. Additionally, it used a randomized technique to control $fw$, promoting stable movements toward optimal solutions and accelerating convergence. In the FDO, there was a fixed $wf$ of 0 or 1; if $wf$=0, it means the search is stable, and when $wf$=1, it means convergence is faster but explores less. But in IFDO, $wf$ is set to be randomly and dynamically determined within the range of [0, 1], giving a fitness weight close to the target and considering to make it more stable, convergent, and full of coverage, IFDO changes equation 2 to equation 7.

$$fw = \left[ \frac{X_{i,t\,fitness}^*}{X_{i,t\,fitness}} \right] \tag{7}$$

Equation 7 represents the $fw$ if the value of $fw$ is less than or equal to $wf$, then the value of $wf$ is ignored; otherwise, $wf$ adjusts the $fw$ using equation 8, which is a new way to find $fw$ and is avoided by ignoring the $wf$ and fairly participating $wf$ in many cases.

$$fw = fw - wf \tag{8}$$

IFDO has done better than the original FDO and other algorithms in tests that use benchmarks, including those that follow IEEE CEC 2019 standards. It showed better skills in exploring and using information, leading to quicker results and the best solutions. In practical situations, IFDO has found the best solutions in fewer steps, proving it is efficient and strong at finding the best overall solutions.

This work [21] proposed an improved FDO, known as modified of FDO (MFDO), to address the identified challenges and enhance the overall performance of FDO. MFDO has demonstrated superior performance in benchmark tests, including CEC2019 standards, outperforming FDO with faster convergence and better solutions. MFDO optimizes the weight factor ($wf$) by narrowing its range to [0, 0.2], thereby achieving a better balance between the exploration and exploitation phases. Also, MFDO incorporates the sine cardinal ($sin$) function to fine-tune the fitness weight $fw$ and adjust the pace of scout movements, ensuring smoother transitions and improved solution refinement as shown in equation 9.

$$pace = \begin{cases} X_{i,t}^* r^* sinc(\pi * wf) & if\ fw = 0 \\ distance_{best\ bee} * r^* sinc(\pi^* wf) & if\ fw = 1 \end{cases} \tag{9}$$

Meanwhile, $fw$ can be calculated using equation 10:

$$fw = \left[ \frac{X_{i,t\,fitness}^*}{X_{i,t\,fitness}} \right] * sinc(\pi^* wf) \tag{10}$$

The IFDO evolved from the FDO in 2020, incorporating alignment and cohesion behaviors into scout bee movements for enhanced optimization. This paper [22] introduces a further modification called M-IFDO, which replaces alignment and cohesion with Lambda parameters as shown in equation 11, improving computational efficiency and achieving better results than IFDO. M-IFDO tackles two significant limitations of FDO: a reduction in precision when using fewer than five agents and a significant reliance on the number of search agents for effectiveness. M-IFDO lowers the computational burden of IFDO by streamlining the computation process and speeding up convergence. It keeps important features like randomized weight factor adjustments for fitness control. A comparative analysis compares the proposed M-IFDO to five competitive algorithms across most standard test functions, with each competitor excelling in specific cases.

$$X_{i,t+1} = X_{i,t} + pace + Lambda \tag{11}$$

Where, Lambda, a component of the pace, has a value of 0.1. There were two different sets of test functions used by the authors of the FDO and proposed improved FDOs [20, 21, 22]. These were the IEEE CEC 2019 benchmark functions (CEC01, CEC03, CEC04, CEC05, CEC07, CEC08, and CEC10) and the classical benchmark functions (F1, F3, F4, F6, F8, F11, F12, F15, F16, F17, F18, and F19). These benchmark functions are employed to evaluate the effectiveness of optimization algorithms across diverse function types [23-24]. These functions are used to ensure their robustness, selecting both continuous and differentiable functions. The chosen functions vary in complexity, including multimodal functions with local minima, non-separable functions, and high-dimensional problems. They are categorized based on their geometric properties, such as many local minima, bowl-shaped, valley-shaped, steep ridges, and plate-shaped. Moreover, the functions F1, F2, and F3 are characterized by 9-dimensional, 16-dimensional, and 18-dimensional problems, respectively, each containing distinct value ranges. Furthermore, the functions F4-F10 are all 10-dimensional problems with an identical search range of [–100,100]. The functions F4-F10 contain distinct rotation matrices. This diversity in test functions helps assess the robustness of optimization algorithms across different challenges. The studies primarily focused on comparing performance and execution time across these benchmarks; here, we compared the performance FDO with its variants across the test functions, as shown in Table 1.

Table 1: Classical benchmark results of FDO with its variants.

| Test Functions | FDO | | IFDO | | MFDO | | MIFDO | |
|---|---|---|---|---|---|---|---|---|
| | AVG | STD | AVG | STD | AVG | STD | AVG | STD |
| TF1 | 7.47E-21 | 7.26E-19 | 5.38E-24 | 2.74E-23 | 2.62E-59 | 1.41E-58 | 3.44E-24 | 1.12E-23 |
| TF2 | 9.388E-6 | 6.90696E-6 | 0.534345844 | 1.620259633 | 2.52E-28 | 9.98E-28 | 0.5172180 | 0.2712662 |
| TF3 | 8.5522E-7 | 4.39552E-6 | 2.88E-07 | 6.90E-07 | 1.29E-13 | 2.40E-13 | 1.06E-13 | 3.93E-13 |
| TF4 | 6.688E-4 | 0.0024887 | 2.60E-04 | 9.11E-04 | 3.61E-13 | 7.54E-13 | 0.5E-04 | 0.0042682 |
| TF5 | 23.50100 | 59.7883701 | 1.94E+01 | 3.31E+01 | 1.06E+00 | 1.57E+00 | 3.1E+01 | 40.019250 |
| TF6 | 1.422E-18 | 4.7460E-18 | 4.22E+06 | 8.15E-09 | 1.92E-32 | 2.23E-32 | 4.15E+06 | 1099.2393 |
| TF7 | 0.544401 | 0.3151575 | 5.68E-01 | 3.14E-01 | 5.09E-01 | 2.95E-01 | 7.2E-01 | 0.3166318 |
| TF8 | -2285207 | 206684.91 | -2.92E+06 | 2.24E+05 | -3.76E+03 | 4.18E+02 | -3.00E+06 | 148152.30 |
| TF9 | 14.56544 | 5.202232 | 1.35E+01 | 6.66E+00 | 1.95E+00 | 9.91E-01 | 8.979103 | 9.84721 |
| TF10 | 3.996E-15 | 6.3773E-16 | 5.18E-15 | 1.67E-15 | 5.15E-15 | 1.42E-15 | 3.891E-15 | 5.8771E-16 |
| TF11 | 0.568776 | 0.1042672 | 0.525690405 | 8.90E-02 | 6.04E-02 | 3.45E-02 | 0.073453 | 0.039061 |
| TF12 | 19.83835 | 26.374228 | 1.81E+01 | 2.57E+01 | 7.05E-08 | 3.62E-07 | 1.75E+01 | 18.610442 |
| TF13 | 10.2783 | 7.42028 | 4.10E+09 | 1.50E-05 | 3.66E-04 | 1.97E-03 | 4.18E+09 | 3.1299918E7 |
| TF14 | 3.7870E-7 | 6.3193E-7 | 2.68E-07 | 4.68E-07 | 1.63E+00 | 7.46E-01 | 8.5E-07 | 3.4511717E-5 |
| TF15 | 0.001502 | 0.0012431 | 4.03E-16 | 9.25E-16 | 3.07E-04 | 4.87E-19 | 0.002E-16 | 6.7292255E-4 |
| TF16 | 0.006375 | 0.0105688 | 9.14E-16 | 3.61E-16 | -1.03E+00 | 0.00E+00 | 1.94E-16 | 0.0334151 |
| TF17 | 23.82013 | 0.2149425 | 2.38E+01 | 1.24E-01 | -1.01E+00 | 3.07E+01 | 2.20E+01 | 0.3212209 |
| TF18 | 222.9682 | 9.9625E-6 | 2.24E+02 | 2.68E-05 | 3.00E+00 | 4.44E-16 | 2.23E+02 | 0.0133942 |
| TF19 | 22.7801 | 0.0103584 | 3.15E+01 | 1.32E-03 | -3.86E+00 | 2.66E-15 | 3.15E+01 | 0.0789975 |

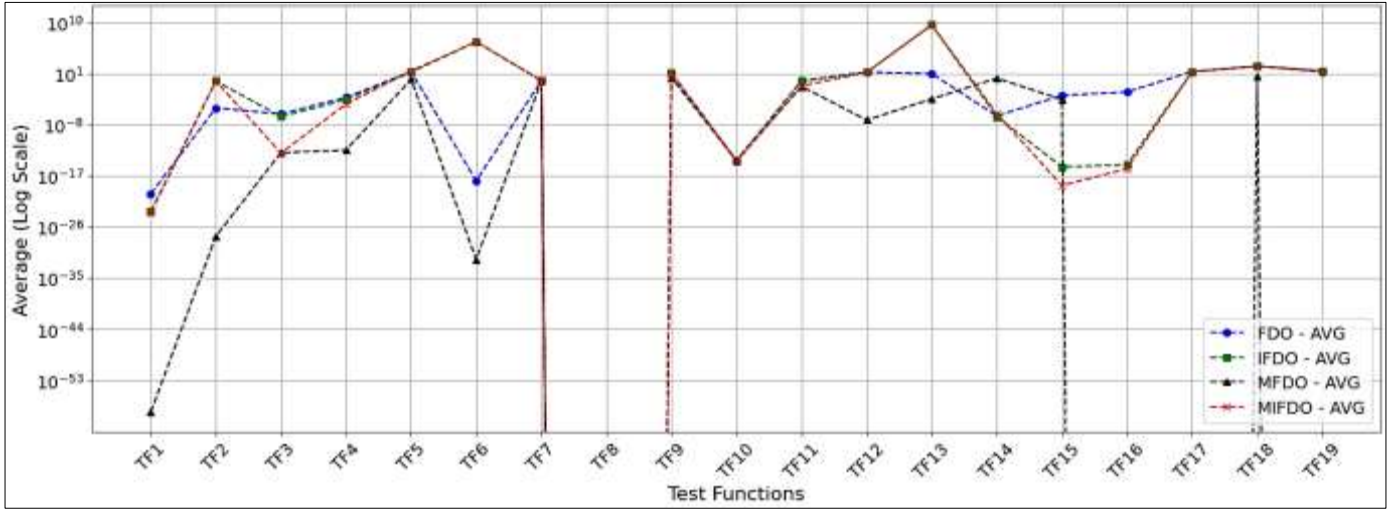

Figure 3. Average (AVG) for FDO with its variants across Test Functions.

The performance of FDO with its variant was evaluated across 19 benchmark test functions for average, as shown in figure 3 and table 1. FDO performed better on simpler tasks, such as TF1 and TF2, with low AVG values, suggesting it is effective at minimizing objective functions in sensitive situations. However, its performance varied on more complex tasks, such as TF7, TF8, and TF9, with larger AVG values indicating difficulty in handling intricate, multi-modal landscapes. Many of the test functions that IFDO did better than FDO, like TF2, TF7, and TF8, had lower AVG values, which means that IFDO found the best solution more quickly when the objective function wasn't too sensitive. IFDO's performance for some challenging functions indicated trade-offs between exploration and exploitation. MFDO had low average values compared to FDO and IFDO but struggled with more complex situations, while MIFDO showed promising results for almost all functions. The findings suggest that algorithm choice should be based on the problem's complexity.

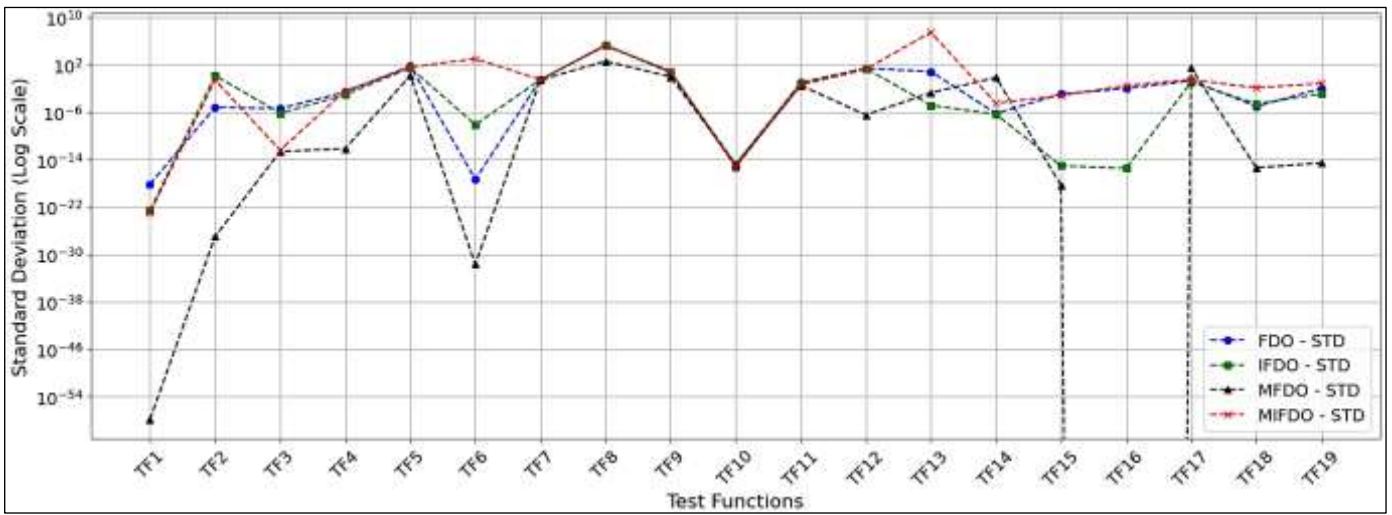

Figure 4. Standard Deviation (STD) for FDO variants with its variants across Test Functions.

The chart in Figure 4 illustrates the results of the standard deviation (STD) of benchmark test functions FDO, IFDO, MIFDO, and MFDO. For FDO, it shows stable results for simple functions such as TF1 and TF2, but its STD values increase for more complex functions, suggesting its reliability has decreased, especially for TF8, which is dynamic and multi-modal. IFDO can reduce some of the variability problems of FDO for many test functions, but it still exhibits high variability for several functions, notably TF5 and TF6. MIFDO observes that certain functions maintain a high degree of consistency in less complex environments, but more complex and variable functions show a more significant performance difference. Nearly all test functions show significant stability in MFDO's results and display substantial variability in more challenging scenarios.

Table 2: IEEE ECE 2019 benchmark results of FDO with its variants.

| Test | FDO | | IFDO | | MFDO | | MIFDO | |
|------|-----|-----|------|-----|------|-----|-------|-----|
| Functions | AVG | STD | AVG | STD | AVG | STD | AVG | STD |
| CEC01 | 4585.27 | 20707.627 | 2651.198672 | 13944.10274 | 4.92E+07 | 4.68E+07 | 2.65E+03 | 1.39E+04 |

| CEC02 | 4.0 | 3.22414E-9 | 4.000002146 | 1.00E-05 | 1.73E+01 | 0.00E+00 | 3.9011 | 0.02113 |
|---|---|---|---|---|---|---|---|---|
| CEC03 | 13.7024 | 1.6490E-11 | 13.70240422 | 4.82E-09 | 1.27E+01 | 8.88E-15 | 13.7024 | 3.50993E-5 |
| CEC04 | 34.0837 | 16.528865 | 31.19516293 | 12.91586061 | 2.82E+01 | 1.52E+01 | 3.12E+01 | 1.29E+01 |
| CEC05 | 2.13924 | 0.085751 | 1.13187643 | 0.070551978 | 1.09E+00 | 4.92E-02 | 1.13E+00 | 7.06E-02 |
| CEC06 | 12.1332 | 0.600237 | 12.12714515 | 0.52079368 | 9.28E+00 | 6.16E-01 | 1.21E+01 | 5.21E-01 |
| CEC07 | 120.4858 | 13.59369 | 115.5677518 | 10.27465902 | 6.00E+01 | 8.95E+01 | 1.36E+01 | 5.79E+02 |
| CEC08 | 6.1021 | 0.756997 | 4.940001939 | 0.891043403 | 4.13E+00 | 4.68E-01 | 4.24E+00 | 8.29E-01 |
| CEC09 | 2.0 | 1.5916E-10 | 2.0 | 3.10E-15 | 2.39E+00 | 3.09E-02 | 2.0 | 5.54501E-4 |
| CEC10 | 2.7182 | 8.8817E-16 | 2.718281828 | 4.44e-16 | 1.63E+01 | 6.66E+00 | 1.91828 | 4.44089E-16 |

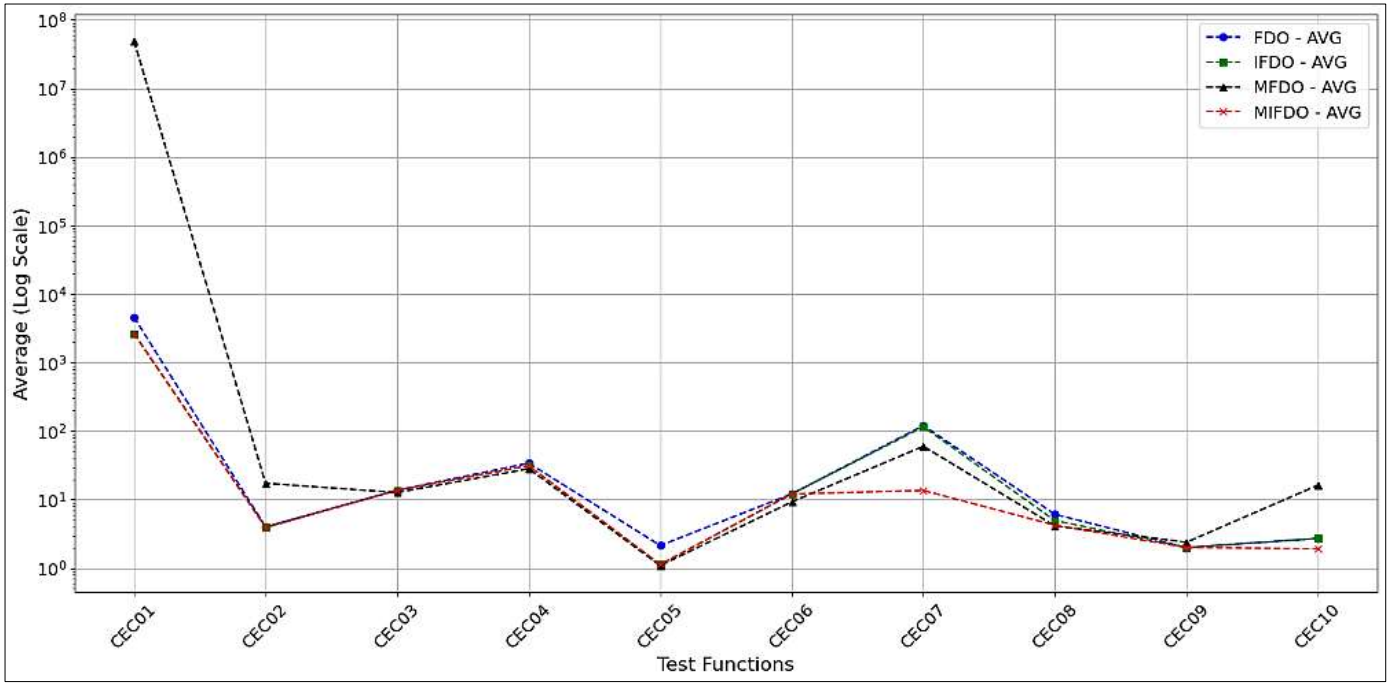

Figure 5. Average (AVG) for FDO with its variants across CES Test Functions.

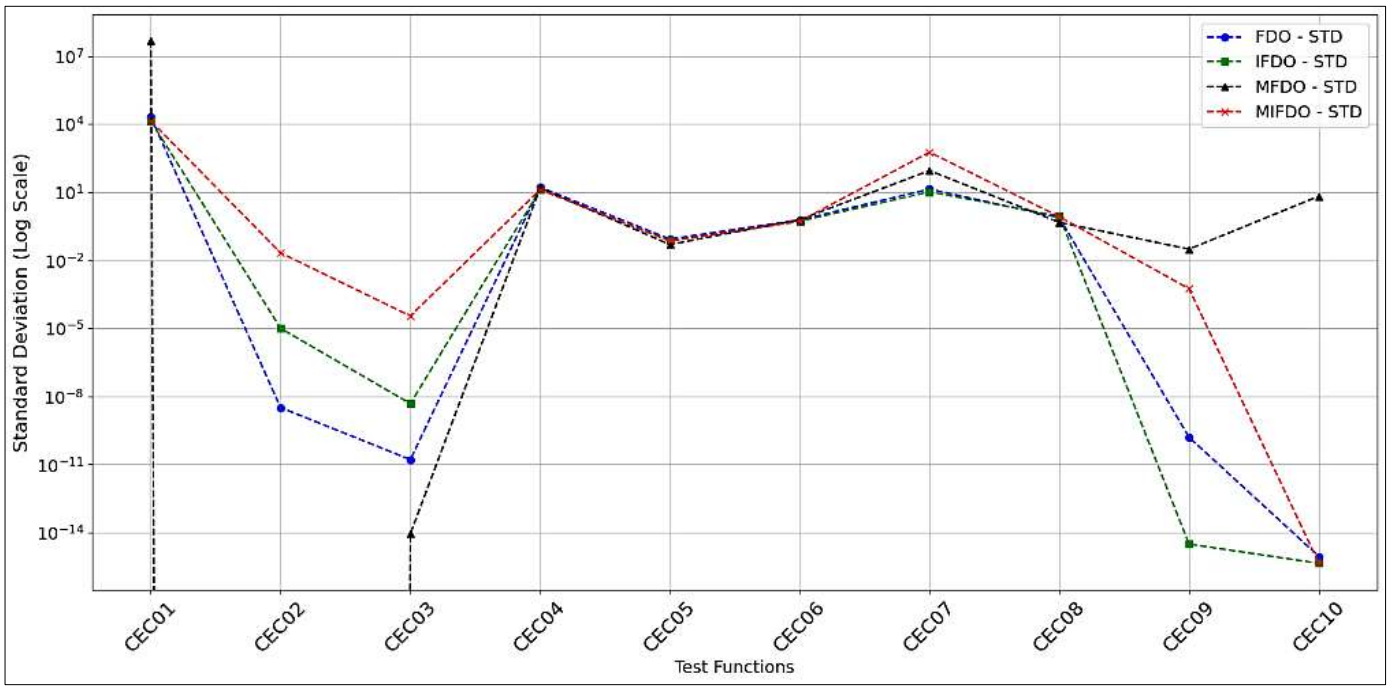

Figure 6. Standard Deviation (STD) for FDO variants with its variants across CES Test Functions.

Table 2 and Figures 5 and 6 display the FDO algorithm and its variants (IFDO, MFDO, and MIFDO). The results show that they perform very differently on the CEC benchmark functions. There is a lot of strength and flexibility in MIFDO. It works better in some test functions like CEC01, where it beats FDO with a lower average (2650.0 vs. 4585.27) and standard deviation (13900 vs. 20707.627). In CEC07, MIFDO demonstrates a significantly lower average (13.6) in comparison to FDO (120.4858), although it exhibits greater variability. MFDO has a lot of potential, but it's not stable, as shown by CEC01 (4.92E+07 AVG, 4.68E+07 STD), which means it's best used in situations where different solutions are helpful. The log-scale plots in figures 5 and 6 correspond with table 2, highlighting MIFDO's consistent performance, FDO's greater variability, and MFDO's unique behavior. The findings underscore the reliability of MIFDO and the potential of MFDO for various optimization challenges.

This paper [25] presents the Adaptive Fitness-Dependent Optimizer (AFDO), a new approach to solving the one-dimensional bin packing problem (1D-BPP), an NP-hard combinatorial optimization problem that is in very high demand for efficient solutions. The association-focused development of optimizers (AFDO) improves the FDO by adding a new First Fit heuristic, the goal is to generate a randomized initial population solution; this ensures diverse and efficient starting solutions. It uses a fitness function to reduce the count of bins and wasted space while employing dynamic adaptations and randomization, enhancing exploration, exploitation, and avoidance of convergence as shown in equations 12 and 13.

$$Minimize \; f(p) = 1 - \frac{\sum_{k=1}^{N}(\frac{fill_k}{C})^k}{N} \tag{12}$$

where $fill_k$ is the amount of weight in bin $k$, and $C$ is the bin's size.

The position of the scout bee is adjusted using a movement rate (pace) that incorporates a positive random number $r$ (ranging from [0, 1]) along with the fitness weight $fw$. This updated equation promotes effective exploration, as shown in equation 13:

$$pace = \begin{cases} X_{i,t} \otimes r & if \; fw = 1 \; or \; 0 \; or \; f(X_{i,t}) = 0 \\ (X_{i,t} \ominus X_{i,t}^*) \otimes fw & if \; 0 < fw < 1 \end{cases} \tag{13}$$

Where these two adaptive parameters, $\otimes$ and $\ominus$, are used to compute the new position of scout bees $X_{i,t}$.

AFDO was tested on well-known benchmark datasets and showed up to 19% improvement in fitness values, and it found solutions with 56% lower execution times compared to previously published algorithms (PSO, CSA, and Jaya), which indicates the capability of AFDO for solving large-scale packing problems., here we compared the performance of AFDO across all three datasets in one table with the PSO, CSA, and Jaya, as shown in Table 3.

Table 3. The results for datasets 1, 2, and 3 (average fitness value).

| Dataset | PSO | | | CSA | | | Jaya | | | AFDO | | |
|---|---|---|---|---|---|---|---|---|---|---|---|---|
| | Max. | Min. | Avg. | Max. | Min. | Avg. | Max. | Min. | Avg. | Max. | Min. | Avg. |
| 1 | 0.228 | 0.199 | 0.2081 | 0.227 | 0.199 | 0.2079 | 0.224 | 0.198 | 0.2053 | 0.224 | 0.198 | 0.205 |
| 1 | 0.203 | 0.194 | 0.1971 | 0.224 | 0.192 | 0.2003 | 0.228 | 0.194 | 0.208 | 0.177 | 0.175 | 0.1759 |
| 1 | 0.209 | 0.174 | 0.1846 | 0.199 | 0.165 | 0.1803 | 0.21 | 0.172 | 0.1872 | 0.178 | 0.142 | 0.1662 |
| 1 | 0.221 | 0.189 | 0.2004 | 0.219 | 0.192 | 0.1988 | 0.218 | 0.191 | 0.1977 | 0.195 | 0.189 | 0.192 |
| 1 | 0.232 | 0.199 | 0.2181 | 0.236 | 0.201 | 0.2258 | 0.232 | 0.199 | 0.2128 | 0.226 | 0.174 | 0.1992 |
| 1 | 0.246 | 0.228 | 0.2405 | 0.266 | 0.227 | 0.2421 | 0.23 | 0.208 | 0.2253 | 0.226 | 0.203 | 0.2095 |
| 1 | 0.213 | 0.175 | 0.2082 | 0.213 | 0.18 | 0.1941 | 0.216 | 0.178 | 0.2085 | 0.211 | 0.148 | 0.1829 |
| 1 | 0.234 | 0.184 | 0.2137 | 0.219 | 0.181 | 0.201 | 0.247 | 0.182 | 0.1993 | 0.192 | 0.16 | 0.1767 |
| 1 | 0.179 | 0.164 | 0.1714 | 0.216 | 0.164 | 0.1803 | 0.248 | 0.129 | 0.1723 | 0.184 | 0.104 | 0.1413 |
| 1 | 0.151 | 0.13 | 0.1383 | 0.151 | 0.114 | 0.1337 | 0.151 | 0.129 | 0.1354 | 0.138 | 0.102 | 0.1187 |
| 1 | 0.26 | 0.194 | 0.2247 | 0.26 | 0.186 | 0.2225 | 0.23 | 0.171 | 0.2112 | 0.214 | 0.158 | 0.1811 |
| 1 | 0.293 | 0.274 | 0.2826 | 0.294 | 0.275 | 0.2836 | 0.307 | 0.282 | 0.2889 | 0.291 | 0.273 | 0.2811 |
| 1 | 0.181 | 0.145 | 0.1701 | 0.174 | 0.145 | 0.1631 | 0.193 | 0.157 | 0.1707 | 0.163 | 0.145 | 0.1524 |
| 1 | 0.266 | 0.253 | 0.2608 | 0.268 | 0.252 | 0.2598 | 0.274 | 0.257 | 0.2615 | 0.264 | 0.252 | 0.2576 |
| 1 | 0.171 | 0.145 | 0.1572 | 0.169 | 0.159 | 0.162 | 0.167 | 0.146 | 0.1548 | 0.158 | 0.141 | 0.1491 |
| 2 | 0.274 | 0.232 | 0.256 | 0.272 | 0.233 | 0.2642 | 0.273 | 0.268 | 0.2707 | 0.248 | 0.211 | 0.2297 |
| 2 | 0.195 | 0.138 | 0.1797 | 0.184 | 0.179 | 0.1819 | 0.186 | 0.18 | 0.1836 | 0.158 | 0.111 | 0.1484 |
| 2 | 0.199 | 0.151 | 0.169 | 0.158 | 0.15 | 0.1536 | 0.193 | 0.152 | 0.1667 | 0.131 | 0.108 | 0.1229 |
| 2 | 0.129 | 0.113 | 0.1205 | 0.128 | 0.117 | 0.1219 | 0.129 | 0.118 | 0.1236 | 0.122 | 0.113 | 0.118 |
| 2 | 0.061 | 0.06 | 0.0605 | 0.196 | 0.06 | 0.074 | 0.195 | 0.06 | 0.0738 | 0.061 | 0.06 | 0.0602 |

| 2 | 0.177 | 0.129 | 0.150364 | 0.18 | 0.125 | 0.1507 | 0.175 | 0.127 | 0.1473 | 0.149 | 0.11 | 0.1222 |
|---|-------|-------|----------|------|-------|--------|-------|-------|--------|-------|------|--------|
| 2 | 0.076 | 0.076 | 0.076 | 0.086 | 0.076 | 0.077 | 0.08 | 0.076 | 0.0764 | 0.076 | 0.075 | 0.0758 |
| 2 | 0.085 | 0.065 | 0.0759 | 0.085 | 0.066 | 0.0724 | 0.082 | 0.069 | 0.073 | 0.073 | 0.059 | 0.0646 |
| 2 | 0.065 | 0.062 | 0.0635 | 0.065 | 0.062 | 0.0633 | 0.064 | 0.06 | 0.0629 | 0.064 | 0.059 | 0.0624 |
| 2 | 0.056 | 0.051 | 0.0534 | 0.053 | 0.051 | 0.0519 | 0.056 | 0.052 | 0.053 | 0.05 | 0.048 | 0.0482 |
| 3 | 0.2 | 0.173 | 0.186 | 0.197 | 0.182 | 0.1903 | 0.2 | 0.187 | 0.1958 | 0.185 | 0.171 | 0.1761 |
| 3 | 0.199 | 0.179 | 0.187 | 0.198 | 0.181 | 0.1871 | 0.192 | 0.18 | 0.1845 | 0.182 | 0.171 | 0.1732 |
| 3 | 0.202 | 0.186 | 0.1939 | 0.201 | 0.182 | 0.1934 | 0.195 | 0.192 | 0.1941 | 0.182 | 0.181 | 0.1814 |
| 3 | 0.2 | 0.188 | 0.1923 | 0.196 | 0.187 | 0.1902 | 0.196 | 0.186 | 0.1916 | 0.186 | 0.174 | 0.1816 |
| 3 | 0.191 | 0.165 | 0.1771 | 0.186 | 0.165 | 0.1762 | 0.187 | 0.166 | 0.1787 | 0.173 | 0.158 | 0.1655 |

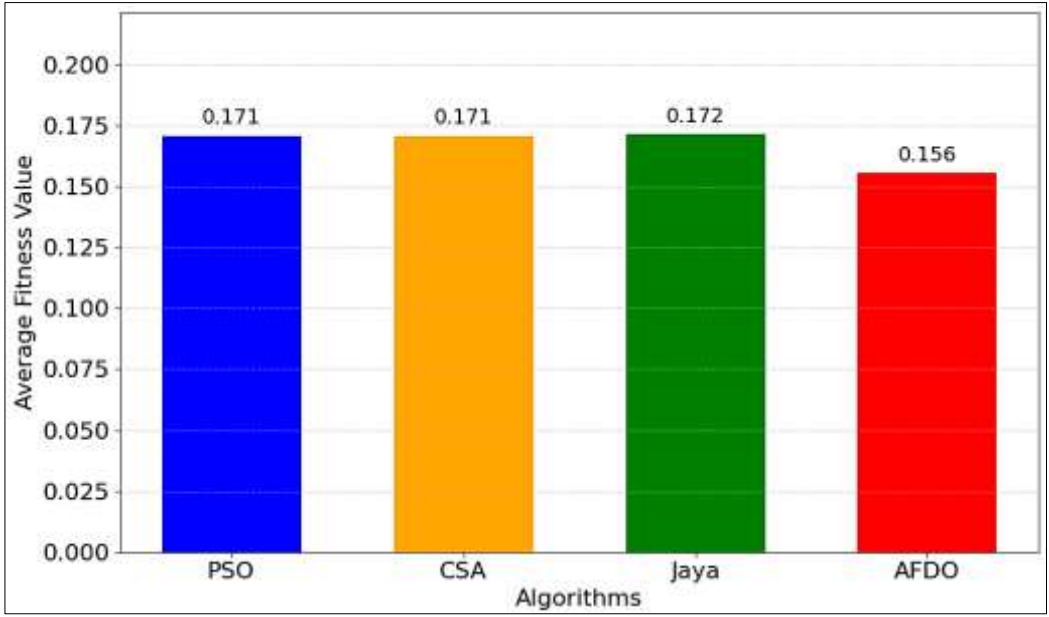

Figure 7. Average Performance of Algorithms Across Datasets.

As shown in table 3 and figure 7. the effectiveness of four algorithms—PSO, CSA, Jaya, and AFDO—was evaluated across three datasets. The means for each algorithm were determined by averaging the fitness scores across all three datasets. Among the algorithms, AFDO showed the best results with the lowest average fitness value, succeeded by Jaya. CSA and PSO had similar average fitness values, suggesting a relatively diminished performance efficiency. This comparison highlights AFDO's uniformity across datasets, resulting in the most advantageous fitness outcomes.

In the paper [26], the authors proposed Improved FDO as an enhancement of the original FDO to better address the economic load dispatch (ELD) problem. The proposed FDO has important improvements like a faster convergence rate, better weighting functions, and updated pacing mechanisms that make it better at solving difficult, high-dimensional problems without achieving premature convergence. According to figure 8 [26], the proposed FDO did better than the original FDO in a standard 24-unit system with changing power demands. It had lower transmission losses, lower fuel costs, and better emission allocations. The authors introduced new population initialization techniques using a quasi-random Sobol' sequence which termed as $Sobol\ [x, y]$, which can be generated over the nonlinear approximation of $S^d$, where $S^d$ is the hypercube with the interval [0,1]as shown in equation 14 to improve exploration of the solution space.

$$\lim_{x \to \infty} \frac{1}{x} \sum_{i=1}^{x} f(S_i) = \int_{S^d}^{i} f. \tag{14}$$

This paper updates the weight factor using the chaotic-sine-map to maintain balance control convergence. When the weight is too low or height, the sine wave helps keep it stale and balanced during the process, as shown in below equation 15.

$$S_{map} = \frac{m}{4} \sin(\pi x_i) \tag{15}$$

Here, $m$ is the controlling factor; the range is $0 < m < 4$, where the authors put 0.3 to $m$ with the most suitable sequence, and in relation to the weight factor, it transforms into this below equation 16.

$$w_s = \frac{m}{4} \sin(\pi w f) \tag{16}$$

In this study, the authors used ANOVA statistical analysis to verify these results, thereby reinforcing the enhanced capabilities of the algorithm. Still, they observed that while proposed FDO often outperforms the fundamental FDO, it requires more parameter adjustment, introducing extra complexity to its execution.

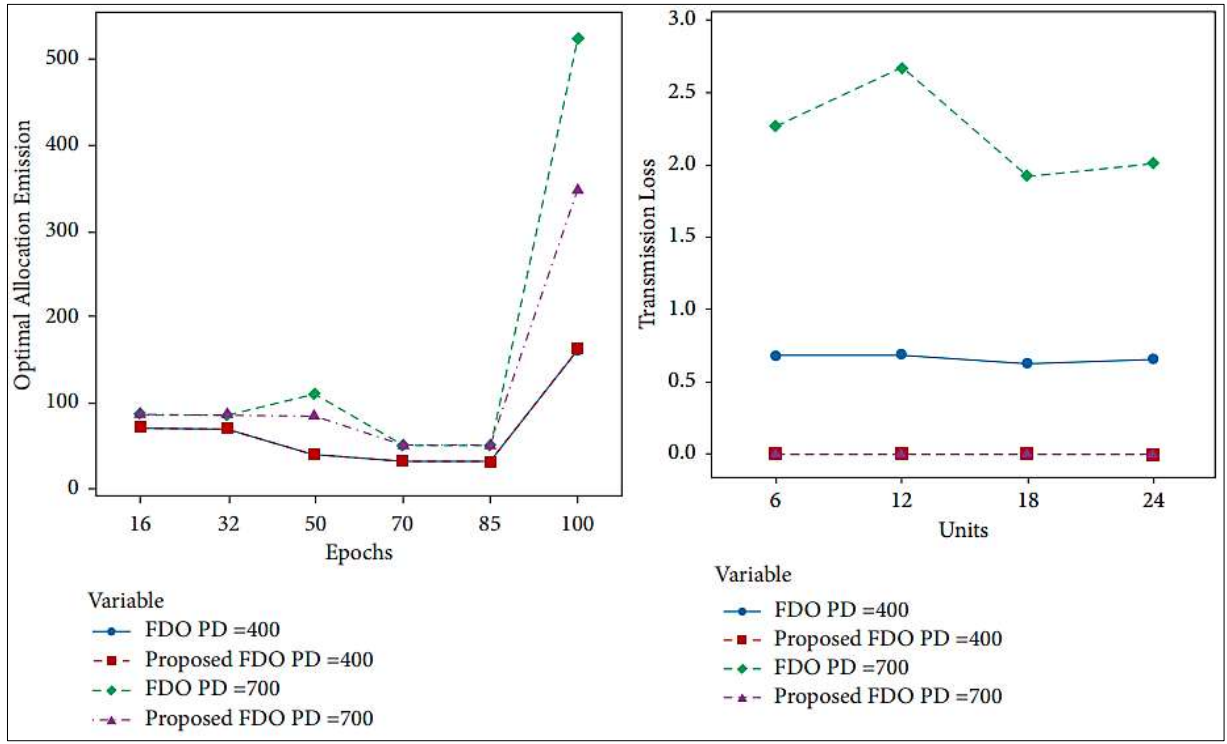

Figure 8 [26].

(A) Convergence of Emission Allocation.          (B) Transmission Loss Comparison.

The result is shown in Figure 8, where the performance of the enhanced FDO over its original version is highlighted. It shows the convergence trend for the optimal emission allocation at power demands of 400 MW and 700 MW. The enhanced FDO achieves faster convergence to an optimal solution, more uniformly, particularly in the early iterations. Also, the loss of transmission for 24 thermal units is included in the figure, showing that the proposed FDO mitigates loss better than the original one. These numbers highlight the proposed FDO's greater efficiency in attaining improved energy optimization outcomes.

### 3.2 Chaotic FDO

This paper proposes the Chaotic Fitness Dependent Optimizer (CFDO) by embedding chaos theory in the FDO [27]. Applying chaos to optimization algorithms enables them to quickly escape local optima, resulting in a more convenient and rapid convergence status. The commonly used algorithm for the chaotic map initializes populations and updates positions using random numbers. The authors modified the original FDO by integrating chaotic maps, boosting the algorithm's diversity and exploratory capabilities. A random variable, denoted as $r$, is used to update the pace vector that defines the position of new solutions influenced by chaotic dynamics. To further improve robustness, a technique was introduced to adjust the bee population when they stray outside the defined search space. The CFDO's performance was evaluated using the CEC2019 benchmark functions as shown in table 4, demonstrating significant improvements over the original FDO and outperforming algorithms such as GA and Chaotic Swarm Optimization (CSO).

Table 4. Comparison Results of WOA, FDO, GWO, and CFDO on CEC2019.

| F | CSO | | GA | | PSO | | CFDO | | FDO | |
|---|---|---|---|---|---|---|---|---|---|---|
| | Avg. | STD | Avg. | STD | Avg. | STD | Avg. | STD | Avg. | STD |
| 1 | 3.66E+0 9 | 3.55E+0 9 | 5.32E+0 4 | 7.04E+0 4 | 1.47127E+1 2 | 1.32362E+1 2 | 1.39E+1 1 | 3.20E+1 1 | 5.56E+1 0 | 4.80E+1 0 |
| 2 | 19.5388 6 | 0.60850 8 | 17.3502 | 17.3491 | 1.52E+04 | 3.73E+03 | 1.90E+0 1 | 6.69E+0 0 | 2.30E+0 1 | 2.08E+0 1 |
| 3 | 13.7024 1 | 8.33E- 06 | 12.7024 | 13.7024 | 1.27E+01 | 9.03E-15 | 1.27E+0 1 | 6.82E- 04 | 1.27E+0 1 | 5.33E- 15 |

| 4 | 198.9105 | 81.32489 | 6.23E+04 | 61986.61 | 1.68E+01 | 8.20E+00 | 2.82E+02 | 3.48E+02 | 1.18E+02 | 4.15E+01 |
|---|---|---|---|---|---|---|---|---|---|---|
| 5 | 2.753796 | 0.192018 | 7.5396 | 7.2765 | 1.14E+00 | 8.94E-02 | 1.76E+00 | 7.39E-01 | 1.33E+00 | 1.93E-01 |
| 6 | 11.66279 | 0.732559 | 7.4005 | 6.6877 | 9.31E+00 | 1.69E+00 | 1.04E+01 | 1.99E+00 | 1.25E+01 | 8.41E-01 |
| 7 | 457.0046 | 141.4665 | 791.742 | 697.896 | 1.61E+02 | 1.04E+02 | 6.43E+02 | 3.82E+02 | 7.17E+02 | 2.62E+02 |
| 8 | 5.679993 | 0.47298 | 6.1004 | 5.8228 | 5.22E+00 | 7.87E-01 | 6.04E+00 | 4.83E-01 | 6.21E+00 | 5.26E-01 |
| 9 | 15.06303 | 11.49835 | 5.31E+03 | 5.29E+03 | 2.37E+00 | 1.84E-02 | 4.27E+00 | 6.07E-01 | 4.57E+00 | 7.56E-01 |
| 10 | 21.40961 | 0.087453 | 20.1059 | 20.0236 | 2.03E+01 | 1.29E-01 | 2.00E+01 | 2.15E-02 | 2.04E+01 | 1.99E-01 |

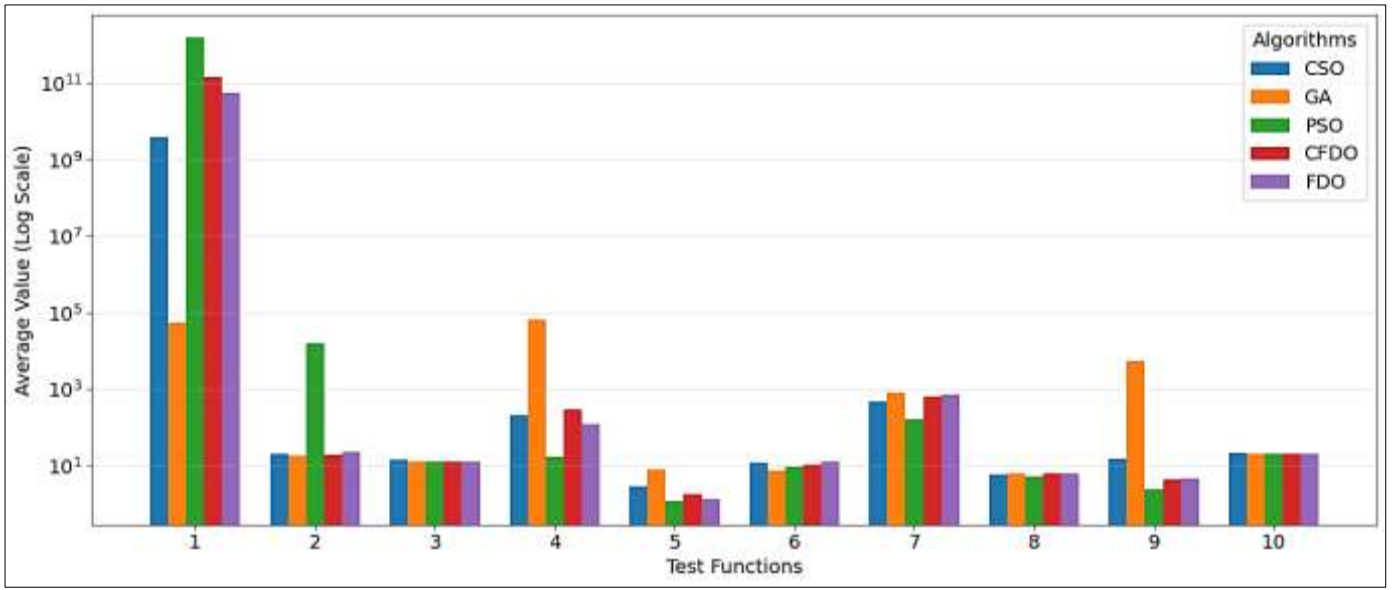

Figure 9. Average Values Across Algorithms.

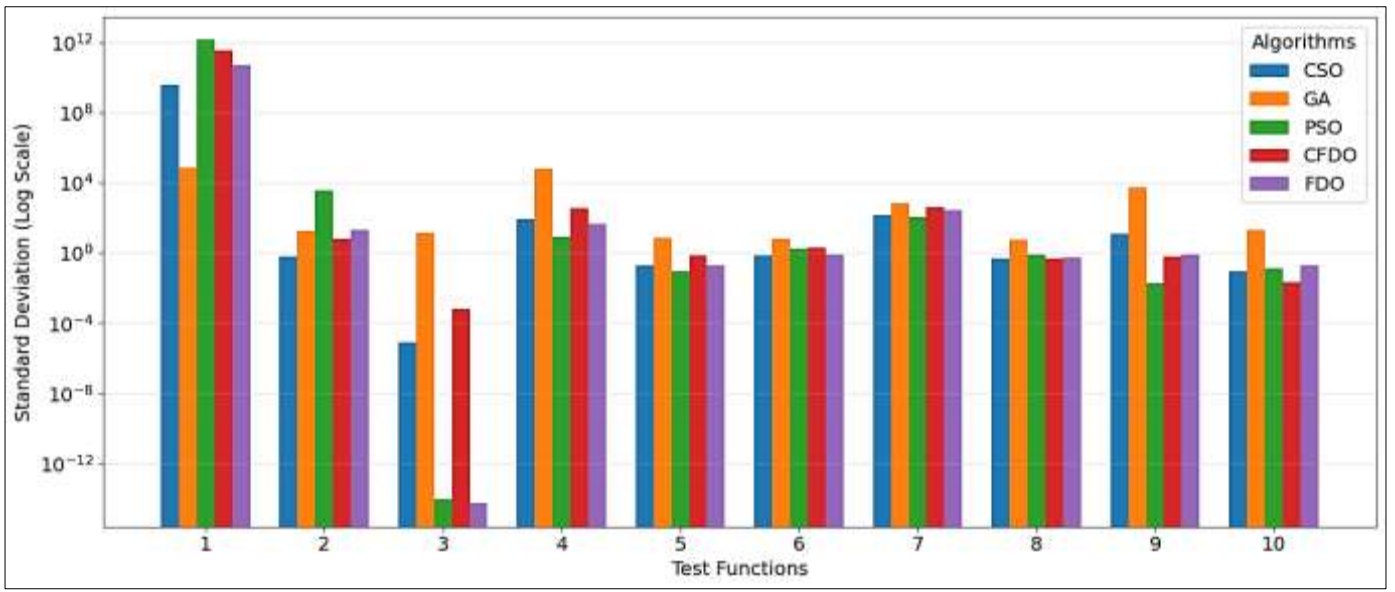

Figure 10. Standard Deviations Across Algorithms.

From the table 4 and image 9 and 10 the CFDO shows enhanced performance relative to multiple standard optimization techniques when assessed using the CEC2019 benchmark suite. The comparison analysis indicates that CFDO exceeds GA, CSO, and FDO regarding solution quality on the majority of benchmark functions. Among the chaotic maps evaluated in CFDO, the Singer map has the most favorable outcomes, whereas the tent map shows the least efficiency. Although PSO demonstrates competitive performance and occasionally outperforms CFDO, the overall findings underscore CFDO's strong proficiency in complex optimization tasks. The data highlights that CFDO is a strong rival in optimization, consistently producing superior outcomes across diverse test functions.

## 3.3 Hybrid FDO

Integrating FDO with other algorithms increases convergence speed, maintains a balance between exploration and exploitation, and enhances solution quality. This hybridization resolves the limitations of FDO, making it more robust for handling complex optimization problems [28], [29].

In this paper [28], the hybrid SCA-FDO (hSC-FDO) optimizes a Fractional Order Integral-Tilt Derivative with Filter (FOI-TDN) controller for load frequency control in power systems with renewable energy sources by combining FDO and the Sine-Cosine Algorithm (SCA). The hybridization leverages SCA's ability to refine local solutions and FDO's global exploration capability, enabling robust performance under practical constraints such as communication delays and generation rate limits. Compared to standalone FDO, PSO, and Firefly Algorithm (FA), the results showed big improvements. Overshoot, undershoot, and settling times were shortened, and stability was improved by using advanced controllers and energy storage systems. This paper [29] hybridized FDO with Bernstein polynomials (BPs) to solve nonlinear optical control problems (NOCPs). The polynomials transformed the original problems into error minimization tasks, while FDO optimized the polynomial coefficients. This hybrid method provided accurate and efficient solutions, achieving lower absolute errors compared to conventional techniques. It proved particularly effective for complex dynamic systems, showcasing improved reliability, stability, and flexibility in numerical solution generation. This work [30] incorporated FDO into the Ant Nesting Algorithm (ANA), a metaheuristic inspired by the behaviors ants use while building nests. This hybrid approach combined FDO's global search abilities with ANA's local enhancement strategies to balance exploration and exploitation. The hybrid algorithm did better than several cutting-edge optimization methods, like PSO, GA, and Whale Optimization Algorithm (WOA), when used to 26 benchmark functions and engineering problems, such as antenna array design. It gave better solution quality and faster convergence. Overall, the hybridization of FDO enhanced the efficiency and versatility of optimization techniques, making them more applicable for real-world problem-solving.

This paper [31] focuses on enhancing automatic generation control (AGC) in interconnected power systems with multiple sources, for example, hydro, gas, and thermal units. The researchers employed FDO to enhance the performance of a modified controller structure, known as Integral-Proportional-Derivative (I-PD), by reducing overshoot, settling time, and enhancing robustness. FDO's ability to balance exploration and exploitation allowed for effective parameter tuning for the controllers. The FDO-optimized controller did better than traditional methods like PSO and Firefly Algorithm when tested against nonlinearities like Generation Rate Constraint (GRC), Boiler Dynamics (BD), and Time Delay (TD). It had a better dynamic response, faster convergence, and lower error metrics, like the Integral Time-Weighted Square Error (ITSE). This work [32] focused on using FDO in machine learning applications, specifically for optimizing neural networks in fault detection tasks. The study used FDO to get the best weights and biases in Multi-Layer Perceptron (MLP) and Cascade MLPs (CMLP), which led to very accurate classification. The FDO-optimized models consistently attained 100% accuracy across all tested datasets, outperforming other optimization techniques such as Grey Wolf Optimizer (GWO). While FDO required slightly longer runtimes, its ability to effectively explore and exploit solution spaces made it superior for fault detection in industrial contexts. The study highlighted FDO's strength in handling small datasets and its reliability in predictive diagnostics. This paper [33] hybrid Sine Cosine and FDO (SC-FDO) has upgraded the FDO. It improves an exploitation ability and convergence speed. The enhancement on pace-updating strategy enhances the exploitation of the algorithm, enabling it to search locally and escaping from local optimum. An addition of a random weight factor and additional changes achieves better performance in terms of exploration versus exploitation balance. The SC-FDO exceeds the performance of the original FDO and other much-studied optimization algorithms on 29 benchmark test functions, providing optimal or near-optimal solutions some. Wilcoxon rank-sum test confirms that it outperforms them. SC-FDO can also handle time series datasets that contain missing data imputation challenges.

In order to improve Maximum Power Point Tracking (MPPT) in thermoelectric generator (TEG) systems under dynamic conditions, this work [34] integrates FDO with Generalized Regression Neural Networks (GRNN) and creates a novel approach known as GRNNFDO. This approach leverages swarm intelligence, a fitness function, and a pace variable to enhance the exploration and exploitation capabilities of the algorithm. The GRNNFDO method demonstrates significant advancements, including improved tracking efficiency exceeding 99%, reduced oscillations in the output, and enhanced adaptability to varying conditions. Additionally, it achieves a notable reduction in tracking time, ensuring rapid and accurate performance. Statistical tests show that GRNNFDO works better than older methods, especially when looking at relative error, mean absolute error (MAE), and root mean square error (RMSE). These improvements make GRNNFDO a promising and robust solution for efficient MPPT in TEG systems, even in fluctuating environments. In this work, the paper [35] FDO has been extended by its integration with the machine learning technique to enhance COVID-19 patient classification based on clinical data. The optimization binds the fitness weight $fw$ within the range [0, 1], and avoids the selection of zero values that could result in division-by-zero errors. The performance of five different models on three datasets was presented by this researcher. The results indicated that FDO significantly outperformed other machine learning models in context. Though the FDO algorithm takes longer runtime compared with other algorithms like GWO, but it has higher accuracy for the classification of COVID-19 patients and could be an important tool for early diagnosis and intervention in clinical settings. This work [36] hybridizes FDO with BPs and GA to create modified versions of FDO, known as FDO-BP and GA-BP, respectively. This is seen as an additional smart hybridization method that can improve the optimization process in NOCPs by combining evolutionary algorithms with techniques for approximation. The authors tried to reduce the absolute error values along with increasing the robustness of the solution. The study demonstrates that the proposed hybrid scheme can obtain the best solutions with significantly smaller errors compared to other existing numerical approaches. Comparative analysis was performed to obtain schemes using FDO-BP and GA-BP, which had good performance in convergence speed and solution quality when applied to different NOCPs; this validated the efficiency of the proposed approach.

Table 5: Summary of FDO Variant Applications

| Reference | Year | Key Improvement | Advantages | Disadvantages | Metrics | Applications |
|---|---|---|---|---|---|---|
| [20] | 2020 | Enhanced exploration-exploitation balance with alignment and cohesion behaviors. | Faster convergence, improved stability, and better solutions. | Requires additional parameter tuning. | IFDO compares with FDO, GA, PSO, and DA in benchmarks and real-world applications, demonstrating better exploration, avoidance of local optima, and convergence, which confirms its efficacy on classical and CEC 2019 benchmarks. | Aperiodic antenna array design and pedestrian evacuation models |
| [21] | 2022 | Narrowed weight factor range; use of sinc function for better movement control. | Faster convergence, smoother transitions. | Performance depends on weight factor adjustment. | MFDO compared to GWO, ChOA, GA, and BOA using CEC2005 and CEC2019 benchmarks and MFDO against FDO, IFDO, SC-FDO, and CFDO using 19 classical benchmark functions, achieving better results in multiple test functions and improving FDO's | Job scheduling, Antenna Array Design, and Traveling salesman problem |

| | | | | | performance, particularly in convergence speed. | |
|---|---|---|---|---|---|---|
| [25] | 2020 | Adaptive mechanisms for enhanced exploration and exploitation. | Improved efficiency in bin packing problems. | Focused on specific problem types. | | One-dimensional bin packing problems (1D-BPP). |
| [27] | 2021 | Integration of chaos theory for better randomization. | Escapes local optima, improved exploratory capabilities. | Require tuning of chaotic maps. | The CFDO was assessed through statistical comparison, benchmark functions, and comparison to other optimization algorithms like PSO, GA, CSO, and FDO, focusing on optimization performance and local optima avoidance. | Pressure vessel design and task assignment problems. |
| [29] | 2022 | Hybridizing FDO with Bernstein Polynomials (BPs) to solve Nonlinear Optimal Control Problems with improved accuracy and efficiency. | faster convergence | Computationally intensive and parameter-sensitive | Absolute Error (AE), Performance Index (J), and statistical analysis (Mean, SD) to evaluate accuracy and reliability | solve various Nonlinear Optimal Control Problems |
| [35] | 2023 | FDO hybridized with NNs for classifying COVID-19 patients, improving model training and classification accuracy. | It avoids local optima and models achieve 100% accuracy | Requires more computational time. | Confusion Matrix Metrics and Mean Squared Error (MSE) | Classification of COVID-19. |

## 4. Application of FDO

Many application fields widely use FDO, a flexible algorithm that successfully balances exploration and exploitation. Its applications show its robustness and effectiveness in solving practical problems; for example, it is highly successful in renewable energy and power systems. Various application fields widely adopt FDO. Its deployment shows its management and efficiency in resolving practical problems.

### 4.1 Energy and power system

In the field of energy and power systems, it has shown remarkable success. This study [26] improves the FDO to better solve the economic load dispatch (ELD) problem by lowering fuel costs, emission allocation, and transmission loss. Enhanced FDO uses advanced initialization techniques and dynamic weight factor adjustment with sine maps for improved optimization. It shows big drops in loss of transmission when tested on a 24-unit power system. The study got 7.94E-12, which is the lowest transmission loss possible when the improved fitness-dependent optimizer was used. This means that it was more stable and convergent than the standard FDO (Computational Intelligence). A study by Daraz et al. [28] used a hybrid Sine Cosine algorithm and FDO to improve an FOI-TDN controller for controlling load frequency in a two-area power system. Execution of exploration vs. exploitation was well balanced on FDO. As a result, it outperformed other metaheuristic algorithms such as PSO and FA in terms of controlling overshoot and settling time.

Yet, parametric sensitivity and computational complexity issues are highly demanding. In this study [31], the PID and I-PD controller parameters for automatic generation control in interconnected power systems with multiple sources were optimized using the FDO methodology. These results indicate that the FDO methodology has promising performance improvement capabilities compared to other metaheuristic algorithms such as TLBO, PSO, and FA, particularly when dealing with multiple nonlinearities such as setting time, overshoot, and undershoot. This work [34] proposes a generalized regression neural network hybridized with FDO to investigate performance improvement in centralized TEG systems operating under dynamic and non-uniform temperature conditions. The GRNN-FDO algorithm shows good tracking of global maximum power points, outperforming other algorithms like CSA, PandO, PSO, and GHO, as it results in the fastest tracking time of 110.1 $MS$ while attaining efficiency higher than 99%. It minimizes oscillations around $GMPP$ and increases energy output by 8.3% under $NUTD$ conditions, proving its effectiveness and reliability for MPPT. This work [37] uses a modified fitness-dependent optimizer to determine the optimal placement of DG units in the system to address issues related to voltage fluctuations, instability, and load demand. A proposed method based on the MFDO enhances voltage profiles and reduces power losses, thereby stabilizing them. The results from the IEEE 14-bus and 30-bus systems show a reduction in power loss of up to 62.91% and 64.05%, respectively, leading to an improvement in voltage from 1.001 $p.u.$ to 1.044 $p.u.$, which is significantly faster than the conventional method's 0.955 $p.u.$ to 1.088 $p.u.$

## 4.2 Industrial and Engineering Application

The algorithm FDO has also found success in industrial and engineering applications, in industrial Critical systems, they have used it for fault detection. For example, in this work [20], optimization tasks have been carried out by IFDO, developed based on enhancements to solve the benchmark functions and real-world engineering problems. Testing IFDO against GA, PSO, and WOA has yielded better performances, with the solutions found through IFDO being quicker to converge and more accurate. These have given very good results in practical scenarios such as aperiodic antenna array design and pedestrian evacuation, hence making IFDO suitable for use in engineering optimization problems. This work [27] applies the Chaotic Fitness-Dependent Optimizer to solve pressure vessel design, task assignment, and various other engineering and planning optimization problems. Benchmark functions CEC2019 and CEC2005 have already evaluated the capability of CFDO. It outperforms FDO compared to CSO and GA in all these fields. In real-world applications, this algorithm has demonstrated its superiority over WOA, GWO, and CGWO, particularly in situations where complications may arise, all while maintaining a high degree of accuracy and efficiency. this work [32] presents the application of FDO to train MLP and CMLP models regarding broken versus non-broken steel plate classification. The classification performance of the FDO-based model outperformed other metaheuristics like GWO and MGWO with an overall datasets performance of 100%, despite a slight increase in computational runtime. Thus, FDO can be applied for fault detection in steel plates during its early stage for safety and reliability.

## 4.3 Healthcare Application

In the healthcare sector, some researchers have applied FDO to healthcare issues, for instance, in this paper [18] FDO is used in IoT-based healthcare systems for data aggregation, prediction, and segmentation. It efficiently processes sensor and medical device data, addressing power consumption issues. In a case study, FDO achieved a global best fitness of 0 in just 2 iterations, outperforming other algorithms like GA, PSO, SSA, DA, and WOA in fitness solutions and convergence speeds, demonstrating its high effectiveness in IoT healthcare applications. This work [35] uses FDO models and neural networks to classify COVID-19 patients as positive or negative, using three datasets with demographic and clinical data. FDO-based models achieved 100% accuracy, outperforming GWO and Modified GWO (MGWO). However, FDO required more runtime than other algorithms, highlighting a trade-off between accuracy and computational efficiency.

## 4.4 Other application

The FDO has been applied to a variety of fields, for example, in numerical optimization, the FDO and Bernstein polynomials (BPs) are combined in the study [29] to solve nonlinear optical control problems (NOCPs). FDO was selected because it can effectively identify global solutions without derivatives and manage nonlinear difficulties. According to the results, the hybrid FDO-BP approach performed noticeably better than previously developed approaches since it reduced absolute errors in state variables, control variables, and the performance index and achieved higher accuracy. In the research [33], missing data in a weather dataset was computed using the FDO. The technique enhanced the speed of convergence and the balance between exploration and exploitation by combining FDO with the Sine Cosine Algorithm (SC-FDO). In comparison to FDO and its variations, SC-FDO greatly reduced the computation time and achieved the greatest average accuracy of 90% for imputing missing data at different rates (10%-90%). In education [38], the FDO algorithm trains multi-layer perceptron (MLP) neural networks to predict students' academic outcomes. The FDO-MLP model achieved 97% accuracy, outperforming other methods like backpropagation, FDO-CMLP, and GWO. This better accuracy, faster convergence, and better local optima avoidance show that it works well for educational data classification tasks, getting around problems like slow convergence and local optima stagnation. The study [39] improves the Social Force Model (SFM) for mass evacuation by incorporating WOABAT-IFDO (Whale-Bat and Improved Fitness-Dependent Optimization) to enhance evacuation time and decrease congestion. The optimization technique strategically distributes guide indicators to efficiently lead agents toward exits, even in obstructions and fluctuating crowd numbers. Statistical analysis and evacuation guidelines show that the optimized model disperses clogging behavior and produces better evacuation times than the traditional SFM.

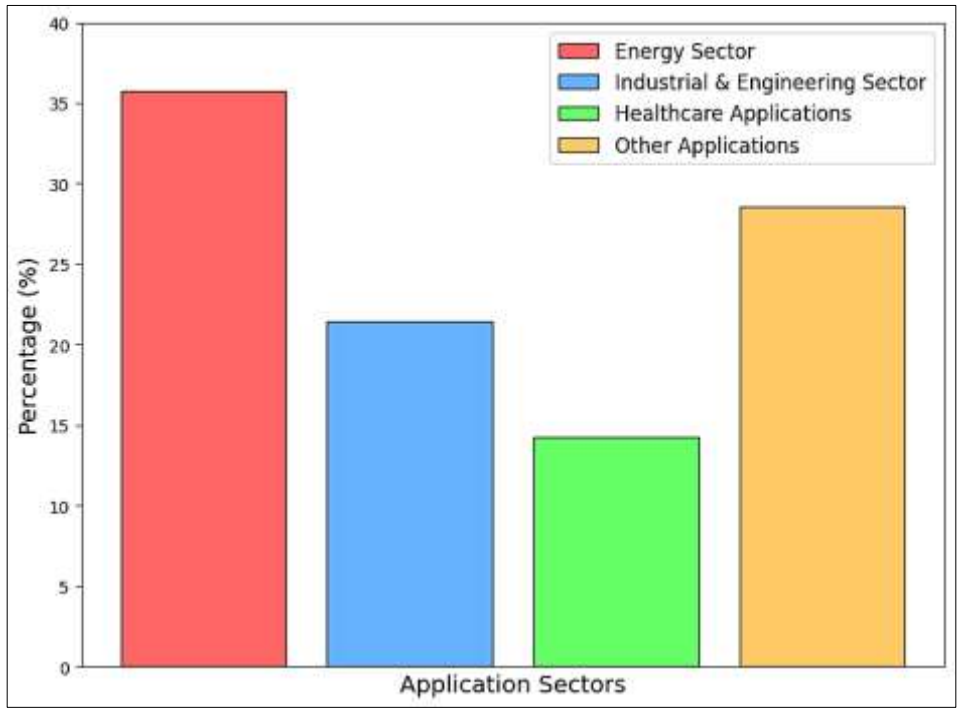

Figure 11. Distribution of FDO Applications by Sector.

Figure 3 highlights how FDO applications are distributed across many domains, with the energy sector accounting for 35.73%, industrial sectors accounting for 21.42% of studies, healthcare for 14.28%, and other applications for 28.57%. This highlights the algorithm's adaptability and wide-ranging influence across multiple areas.

## 5. Future Work

FDO, as the optimization algorithm, serves a crucial role in solving real-world challenges such as engineering designs and machine learning. Despite its success, there are still issues with global convergence speed, dealing with high-dimensional search spaces, and finding a solution. But by using one of the advanced techniques or all of them, such as advanced initialization techniques, dynamic diversity, and better operator selection, we can enhance the robustness of the FDO for a wide range of optimization tasks. This section aims to maintain a practical use focus while reinforcing the theoretical basis of the design through these improvements.

### A. Advanced Initialization for Improved Exploration

The initial setup of scout bees, the agents in FDO, lays the foundation for the algorithm's ability to navigate the search space efficiently. Normally, scout bees are initiated randomly:

$$X_{i,0} \in S, \ i = 1, 2, \dots, N$$

Where:

- $N$: Number of scout bees.

- $X_{i,0}$: Initial position of the i-th scout bee.
- $S$: Search space defined by the problem constraints.

Although this approach is simple, it can lead to coverage of different search areas. This is particularly relevant for high-dimensional or multimodal problems. Therefore, we can propose quasi-random sequences [40], like the Sobol or Halton sequences, to address this problem. These sequences guarantee a more systematic and even distribution of scout bees throughout the search area. For a search space spanning a range $S = [X_{min}, X_{max}]$, the initial position can be initialized as below equation 17:

$$X_{i,0} = X_{min} + (X_{max} - X_{min}) . Q_i \qquad (17)$$

Where $Q_i$ refers to a quasi-random variable particular to the i-th scout bee. This guarantees an optimal distribution of scout bees, increasing global exploration and minimizing the possibility of skipping essential areas within the search space.

## B. Adaptive Diversity Management for Dynamic Control

A key characteristic of FDO is the use of a diverse set of scout flies to maintain a balance between exploration and exploitation. Diversity measures how spread out the scout flies are in the search space. Equation 18 provides the calculation for this:

$$\text{Diversity} = \frac{1}{N} \sum_{1}^{N} \|X_{i,t} - X_{mean}\|^2 \qquad (18)$$

Where the variable $X_{mean}$ represents the average position of all scout bees at iteration $t$, as illustrated in equation 19 below:

$$X_{mean} = \frac{1}{N} \sum_{1}^{N} X_{i,t} \qquad (19)$$

The diversity is necessary, as a balance between exploration and exploitation is critical for the algorithm. A high level of diversity indicates a broad dispersion of scout bees, which would require exploration on a global scale to find new potential regions. Additionally, low diversity signifies the convergence of the algorithm on a solution, indicating a need for greater focus on local exploitation, and vice versa. To successfully manage this transition, the FDO algorithm can apply dynamic thresholds as show in in equation 20 for diversity in the following manner:

$$Threshold_{high(t+1)} = Threshold_{high(t)} - t.\Delta_{high}$$
$$Threshold_{low(t+1)} = Threshold_{low(t)} + t.\Delta_{low} \qquad (20)$$

While high and low thresholds should be initialized in the beginning, using this dynamic adjustment allows the algorithm to modify its focus during the optimization process, first prioritizing exploration and eventually switching to exploitation.

## C. Dynamic Operator Selection for Enhanced Search

As per the measured diversity, the FDO drives the scout bee diversity for the exploration of the search space, where a large value of diversity directs the exploration of unexplored areas, and lower diversity values direct convergence towards optimal solutions. Such an adaptive understanding further enables the FDO to make informed choices between employing efforts in either exploration or exploitation operations, resulting in greater operation efficiency and offering a broadly well-populated solution space.

When diversity is high means ($Diversity > Threshold\_high$), indicating a broad distribution of scout bees, we can use Levy [41] flights to encourage significant random movements. This allows scout bees to explore remote, unexplored areas of the solution space. In this mode, we compute a new position for each scout bee as below in equation 21.

$$X_{i,t+1} = X_{i,t} + levy(\lambda) \qquad (21)$$

Where the Levy flight can be defined by below equation 22:

$$Levy(\lambda) = \frac{\mu}{|v|^{\frac{1}{\gamma}}} \tag{22}$$

Where $\mu \sim N(0, \sigma^2), v \sim N(0, 1)$.

This technique allows scout bees to explore faraway regions, which reduces the possibility of becoming trapped in local optima.

When diversity is low means $(\boldsymbol{Diversity < Threshold\_high})$, scout bees will focus their attention on a specific area. This requires extensive local search. In this case, we can use a chaotic map to guide the refinement process as shown in equation 23.

$$X_{i,t+1} = X_{i,t} + ChaoticFactor(r).(X^* - X_{i,t}) \tag{23}$$

Where $\mathbf{X}^*$, is the best solution and ChaoticFactor(r) could be any function of the chaotic map for example logistic map or Singer map.

Overall, the proposed improvements to the FDO will probably provide significant benefits. Better starting with quasi-random sequences to cover a bigger search space, adaptive diversity management to keep a balance between exploring and exploiting, and better performance by using Levy flights and chaotic maps to solve difficult, high-dimensional, and multimodal problems are some of the suggested improvements. The refinements enhance the algorithm's applicability for multiple real-world optimization tasks, including energy systems and industrial design, therefore providing the FDO with a more robust and adaptable optimization tool.

## 6. Conclusions

In this paper, we present an overview of the FDO, highlighting its origins, variations, and performances in various applications. The swarm behavior of bees inspires the FDO, which has proven to adapt and effectively solve various optimization problems. This study, through a systematic analysis of its variants, such as IFDO, MFDO, MIFDO, and CFDO, underscores the algorithm's strengths, including improved exploration-exploitation balance, faster convergence, and robustness in tackling complex, multimodal problems. These modifications yielded positive results in testing, engineering applications, healthcare, and energy systems, thereby confirming FDO as an essential algorithm for optimization. Regardless, there are limitations to address, particularly with global convergence speed, managing high-dimensional search spaces, and maintaining a balance between exploration and exploitation, we recommended improvements to overcome these limitations, which shows there is still room for improving FDO's performance, and through these techniques, FDO will be able to outperform and be more flexible for real-world problems. To answer the title of this paper is FDO ready for future optimization, indeed we can say yes, because of simplicity, flexibility and handle complex optimization problem, by implementing new advancements such as the development of adaptive and parallel versions and hybridize it with other algorithms. In conclusion, FDO and its variations have achieved substantial advancements in optimization research, obtaining remarkable results across multiple sectors. By overcoming the highlighted challenges and researching the recommended future work, researchers can discover the whole potential of FDO, enabling the development of more robust and efficient optimization solutions.